%% file: sample-base.tex
\documentclass[sigconf]{acmart}
\usepackage{multirow}

\newcommand{\ie}{\textit{i}.\textit{e}.}
\newcommand{\eg}{\textit{e}.\textit{g}.}
\newcommand{\etc}{\textit{etc}.}
\AtBeginDocument{%
  \providecommand\BibTeX{{%
    \normalfont B\kern-0.5em{\scshape i\kern-0.25em b}\kern-0.8em\TeX}}}

\setcopyright{acmcopyright}
\copyrightyear{2022}

\acmConference[Preprint]{}{Madrid, Spain}

\begin{document}


\title{Block-SCL: Blocking Matters for Supervised Contrastive Learning in Product Matching}

\author{Mario Almagro}
\authornotemark[1]
\email{mario.almagro@nielseniq.com}
\affiliation{
  \institution{NielsenIQ - Innovation}
  \city{Madrid}
  \country{Spain}
}

\author{David Jiménez} 
\authornotemark[1]
\email{david.jimenez@nielseniq.com}
\affiliation{
  \institution{NielsenIQ - Innovation}
  \city{Madrid}
  \country{Spain}
}

\author{Diego Ortego} 
\authornotemark[1]
\email{diego.ortego@nielseniq.com}
\affiliation{
  \institution{NielsenIQ - Innovation}
  \city{Madrid}
  \country{Spain}
}

\author{Emilio Almazán}
\email{emilio.almazan@nielseniq.com}
\affiliation{
  \institution{NielsenIQ - Innovation}
  \city{Madrid}
  \country{Spain}
}
\authornote{M. Almagro, D. Jiménez, D. Ortego and E. Almazán contributed equally to this research. E. Martínez contributed to the writing of the document and previous discussions.}

\author{Eva Martínez Garcia}
\email{eva.martinezgarcia@nielseniq.com}
\affiliation{
  \institution{NielsenIQ - Innovation}
  \city{Madrid}
  \country{Spain}
}

\renewcommand{\shortauthors}{ Almagro et al.}

\begin{abstract}
Product matching is a fundamental step for the global understanding of consumer behavior in e-commerce. 
In practice, product matching refers to the task of deciding if two product offers from different data sources (\eg~retailers) represent the same product. 
Standard pipelines use a previous stage called ``blocking'', where for a given product offer a set of potential matching candidates are retrieved based on similar characteristics (\eg~same brand, category, flavor, \etc). 
From these similar product candidates, those that are not a match can be considered hard negatives. 
We present \textbf{Block-SCL}, a strategy that uses the blocking output to make the most of Supervised Contrastive Learning (SCL). 
Concretely, Block-SCL builds enriched batches using the hard-negatives samples obtained in  the blocking stage. 
These batches provide a strong training signal leading the model to learn more meaningful sentence embeddings for product matching. Experimental results in several public datasets demonstrate that Block-SCL achieves state-of-the-art results despite only using short product titles as input, no data augmentation, and a lighter transformer backbone than competing methods.

\end{abstract}




\maketitle
\section{Introduction}
\label{sec1_intro}
\input{sec1_introduction}

\section{Related work}
\label{sec2_sota}
\input{sec2_relatedWork}

\section{Approach}
\label{sec3_approach}
\input{sec3_approach}

\section{Experiments}
\label{sec4_experiments}
\input{sec4_experiments}

\section{Conclusions}
\label{sec5_conclusions}
\input{sec5_conclusions}

\bibliographystyle{ACM-Reference-Format}
\bibliography{sample-base}


\end{document}

%% file: sec1_introduction.tex
Nowadays e-commerce websites offer hundreds of millions of products as a result of the online expansion of retailers' businesses. 
In this context, performing product matching successfully, \ie~automatically finding offers of the same product from different data sources is key to providing good quality of experience. 
The product matching task is particularly challenging due to a large number of existing products, their high heterogeneity, missing product information, and varying levels of data quality \cite{2020_ACLW_BERT_PM}.
An example of product offers to match from different sources extracted from WDC dataset \cite{2019_WWWC_WDCdataset}: \emph{"Kingston DataTraveler microDuo 3C 32GB Zilver - Prijzen Tweakers"} and \emph{"Kingston Technology GmbH LambdaTek|USB flash drives"}. 
Product matching, then, requires a substantial language understanding and domain-specific knowledge \cite{2020_VLDB_Ditto} to differentiate between positive and negative pairs. We refer the reader to \cite{2018_NAACL_Industry_PMchallenges} for a detailed overview of product matching challenges.

The current trend in product matching is to use the Transformer paradigm \cite{2020_ACLW_BERT_PM, 2020_VLDB_Ditto, 2022_arXiv_SupConLossPM}, which relies on robust pre-trained Transformer models to serve as starting point for solving downstream tasks \cite{2019_NAACL_BERT, 2021_CCL_Roberta}.
For example, the authors in \cite{2020_ISWC_PM_inputConcat, 2020_ICEDT_PMinputConcat, 2020_VLDB_Ditto} concatenate the two product descriptions and subsequently process them using a Transformer encoder, \eg~BERT \cite{2019_NAACL_BERT}, and add a classification head on top to solve product matching as a binary classification task. 

Alternatively, R. Peters and C. Bizer~\cite{2022_arXiv_SupConLossPM} process the product descriptions individually and estimate separate description embeddings, which are then compared in a metric learning fashion and classified as match or not match.  
Metric learning frameworks are very popular in natural language processing (NLP) for a variety of applications including: text classification \cite{2022_arXiv_TextClassificationSCL}, sentence representation \cite{2021_EMNLP_SimCSE}, named entity recognition \cite{2022_ACL_CONTainNER} or text retrieval \cite{2021_ICLR_TextRetrieval}. 
These metric learning methods work by pulling together representations of the same concept and pushing apart representations for different concepts.
Currently, the methods based on Supervised Contrastive Learning (SCL) are leading the progress in metric learning~\cite{2021_NeurIPS_SupContLearn}. SCL proposes a training objective that, within a batch of training samples, considers positives for each sample those of the same class and as negatives the remaining ones. 
This vanilla version of SCL has been adapted or improved in different contexts, \eg~text classification \cite{2022_arXiv_TextClassificationSCL}, product matching \cite{2022_arXiv_SupConLossPM}, learning with label noise \cite{2021_CVPR_MOIT} or long-tailed classification \cite{2021_ICCV_PaCo}. 
In particular, authors in \cite{2022_arXiv_SupConLossPM} use vanilla-SCL for product matching, where positive/negative pairs are simply positive/negative pairs of product offers. 
We adopt a similar approach and focus on a key aspect overlooked in \cite{2022_arXiv_SupConLossPM}: the hard negatives.

Product matching usually involves a blocking pre-processing step to discard entries that are unlikely to be matches, thus reducing the matching problem to the resulting blocking entries. 
Moreover, hard negatives play a key role in achieving better similarity learning \cite{2017_ICCV_SamplingMatterMetlearn, 2020_ECCV_HardMining}, thus we believe that the blocking is a natural source of hard negatives that can boost metric learning for product matching.

In this paper, we propose to use blocking information to enhance supervised contrastive learning for product matching in e-commerce. The main contributions of the paper are listed below:

\begin{itemize}
    \item We demonstrate that considering hard negatives from the blocking information during the batched optimization benefits the learning process and boosts product matching performance.
    \item We evaluate our approach in 6 public datasets comparing results with top-performing methods. Moreover, in the ablation study, we show the impact of adding more positive/ negative samples in the blocking.
    \item We demonstrate that solely using hard negatives for SCL without using larger backbones and tricks from related methods obtains the best overall results while improving the efficiency of the models considerably.
\end{itemize}

%% file: sec2_relatedWork.tex
The problem of product matching refers to the task of finding product offers in a large dataset given a textual query for product offers. The main challenge lies in the fact that query and candidate offers come from different domains \eg~ retailers.
In practice, this problem is approached with a two-stage pipeline: 
\begin{enumerate}
    \item Blocking: a light comparison on the entire dataset is conducted with the goal of identifying a subset of descriptions that are potential matches for the product query. It is expected to obtain a subset without false negatives.
    \item Matching: the descriptions from the ``blocking'' stage are classified as match or not match for the input query.
\end{enumerate}
Most works found in the literature deal only with the second stage in isolation. In particular, the problem is treated as a pair classification problem, where each pair is independent of the other. However, we believe that there is relevant information in the blocking output that we can use to improve the matching stage.

Early solutions for product matching were based on hand-crafted features and rules that provided a high level of interpretability at the cost of involving human expert knowledge~\cite{2009_VLDB_recordMatchingRules,2017_ICMD_generatingRules}. 
Deepmatcher~\cite{2018_SIGMOD_DeepMatcher} can be considered the first deep learning based model for product matching. 
This method generates independent features for each description using Recurrent Neural Networks, which are then compared and classified as match or not match.

More recently, Transformer-based approaches have shown good performance on solving this task \cite{2020_ICEDT_PMinputConcat, 2020_ISWC_PM_inputConcat, 2020_VLDB_Ditto, 2021_VLDB_jointBERT}. 
These methods usually concatenate pairs of textual product offers and feed them to a large pre-trained language model (\eg~ BERT) that is fine-tuned with a classification head that provides matching/not matching predictions. 

Nevertheless, the performance of Transformer-based models is directly related to the quality of the sentence embeddings used.  
Since BERT-based models do not produce independent single sentence embeddings, they are generated by averaging the outputs of the BERT model or by using the special \textsc{[CLS]} token. 
However, these sentence embedding methods have shown limitations to properly capture sentence semantics \cite{2019_arxiv_BiasesSentenceEncoders,2019_arxiv_BERTscore,2019_arxiv_BertRanking}.
Also, the pre-training task significantly impacts the embeddings quality \cite{2016_ACL_distributedRepresentations}.
Many works leverage the information from BERT models to build sentence embeddings \cite{2019_EMNLP_SBERT, 2020_EMNLP_SentenceEmbForBERT}, and those using self-supervised contrastive learning approaches achieve the state-of-the-art results.
Some of them focus on building positive sentence pairs \cite{2021_EMNLP_SimCSE,2021_arxiv_Consert}, or using prompt learning \cite{2022_arxiv_promptbert,2022_arxiv_Sncse} to reduce token embeddings biases and make the BERT layers capture better sentence semantics that help the network distinguish between soft and hard negative examples.

To the best of our knowledge, there is only one recent work that learns sentence embeddings for product matching via contrastive learning \cite{2022_arXiv_SupConLossPM}. We name this approach  Vanilla-SCL as they follow the original SCL~\cite{2021_NeurIPS_SupContLearn} random strategy for assembling the batches. 
Nevertheless, they ignore the information from the blocking, which is a source of hard negatives. Recent works \cite{2020_ACLW_BERT_PM} have shown the benefits of using hard negatives and positives to obtain better products embeddings. 
We identify the potential benefits of using hard negative examples enclosed in the information from the blocking filtering when applying supervised contrastive learning and propose to include that information during batch construction.


%% file: sec3_approach.tex
The proposed approach relies on two steps: pre-training using SCL with blocking information and product matching classification. First, we train the backbone and projection network to produce robust sentence representations so that offers of the same product are separable from the rest. Then, we freeze the backbone and replace the projection head with a linear layer to perform product matching classification using pairs of product offers.

\subsection{Pre-training}
\label{sec3_1_pretraining}
\begin{figure}[htbp]
    \centering
    \includegraphics{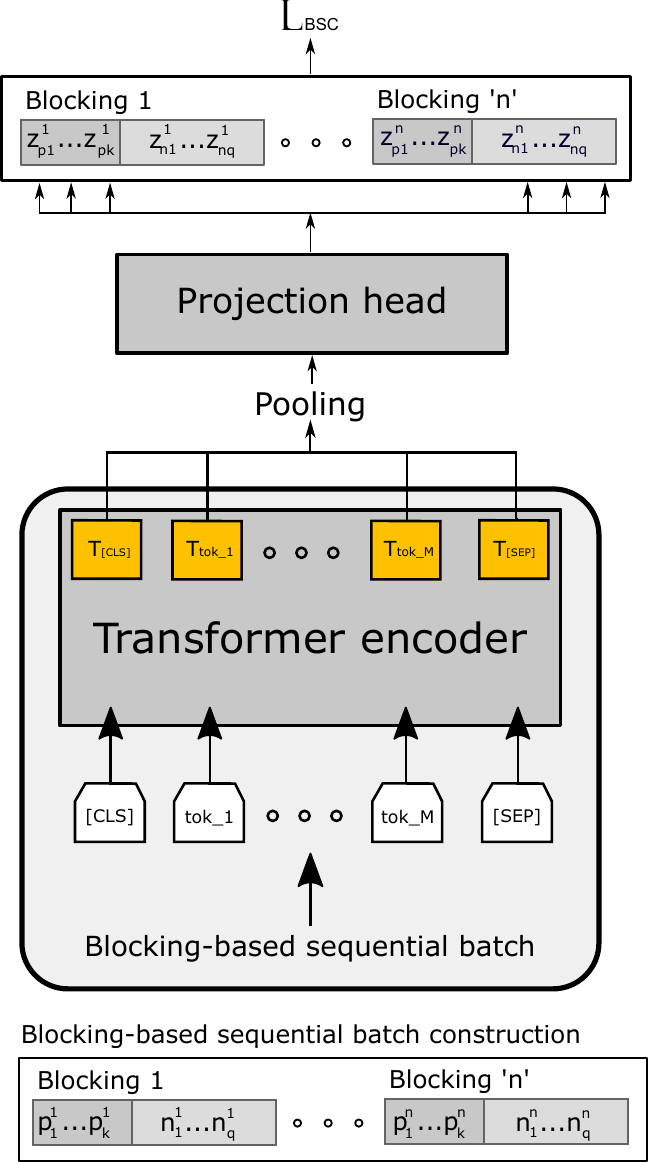}
    \caption{Diagram of the proposed pre-training strategy and the blocking-based batch construction, where 'p' and 'n' refers to positives and negatives offers within the same blocking, respectively.}
    \label{fig:diagram_pretrain}
\end{figure}

Broadly speaking, we adopt the SCL approach from \cite{2021_NeurIPS_SupContLearn} proposed for visual representation learning and adapt it to learn sentence embeddings for product matching. This is achieved by pulling offers of the same product closer in the feature space and pushing away those of other products during training. Figure \ref{fig:diagram_pretrain} presents an overview of the pre-training proposal.

First, we use the sampling strategy depicted in Sec.~\ref{sec3_1_2_batch_construction} to select a batch $\mathcal{B}_{i}$ of product offers $x_{i}$, where $i\in\left[1,\left|\mathcal{B}_{i}\right|\right]$ and $\left|\mathcal{B}_{i}\right|$ is the batch size. The purpose of using blocking information relative to each product is to identify the product offers that should lie closer in the feature space and thus force the model to learn representations that differentiate them.
Every product offer is then mapped to a low-dimensional representation $z_{i}$ by learning a transformer encoder network $f_{\theta}$ and a multi-layer perpectron projection head $g_{\phi}$ with parameters $\theta$ and $\phi$, respectively. 
In particular, the sentence embedding $v_{i}=f_{\theta}\left(x_{i}\right)$ is generated by average pooling of token embeddings produced by $f_{\theta}$. 
Subsequently, the sentence embedding is transformed into a low-dimensional representation $w_{i}=g_{\phi}\left(v_{i}\right)$, which is later $L_{2}$-normalized into $z_{i}=w_{i}/\left\Vert w_{i}\right\Vert _{2}$. Then, the contrastive loss function described in Sec.~\ref{sec3_1_3_scl_loss} is calculated.

\subsubsection{Data preparation}
\label{sec3_1_1_data_preparation}
Most of the datasets in product matching (\eg~WDC \cite{2019_WWWC_WDCdataset}) consist of $m$ pairs of relatively similar product offers labeled to denote if they refer to the same product (match) or not (non-match). Then, each data pair includes a product offer on the left (${\text{left}}_i$), a product offer on the right (${\text{right}}_i$), and a binary matching label $y_i$.
The nature of the data is that matching/non-matching offers correspond to the same/different product, sharing all of them a high degree of textual similarity given that they come from the same blocking. Therefore, we can assume that non-matching pairs provide a source of hard negative samples.

Although product and blocking identifiers are not always provided, such information can be generated using the set of offer pairs. Those pairs sharing some offer would constitute a blocking, while only matching offers would have the same product id. Therefore, we propose to use a match-based graph to assign unique product identifiers to all matching offers, while assigning the same blocking identifier to all non-matching products compared with a specific product.

\subsubsection{Blocking-based batch construction}
\label{sec3_1_2_batch_construction}
Batches are built to contain product offers $x_{i}$ from $n$ different blockings, so offers with the same product id are considered positive samples, and offers not sharing the product id but sharing the blocking id are assumed to be hard negative samples. To do so, data is sampled in such a way that we obtain a set of product ids in every batch for which we have a set $\mathcal{P}_{i}$ of positive offers and a set $\mathcal{N}_{i}$ of hard negative ones. The cardinality of $\left|\mathcal{P}_{i}\right|\leq k$ and $\left|\mathcal{N}_{i}\right|\leq q$, being $k$ and $q$ hyper-parameters that denote the number of positive and negative samples in a blocking to be selected for each product id, \eg~$k=2$ selects $2$ positive offers for the product id associated to sample $x_{i}$. 
 
Note that the sum of positive and negative samples in the batch $\left|\mathcal{P}_{i}\right|+\left|\mathcal{N}_{i}\right|=\left|\mathcal{B}_{i}\right|-1$, where $-1$ denotes the self-contrast case. 
This batch creation strategy forces the network to distinguish between positive and negative pairs that have similar text sequences as they belong to the same blocking, as well as unrelated offers coming from the remaining blockings.

\subsubsection{SCL loss}
\label{sec3_1_3_scl_loss}
The supervised contrastive learning loss is defined as follows:
\begin{equation}
\ensuremath{\mathcal{L}_{SCL}=\frac{1}{\left|\mathcal{B}_{i}\right|}\underset{i\in\mathcal{B}_{i}}{\sum}\frac{1}{\left|\mathcal{P}_{i}\right|}\underset{p\in\mathcal{P}_{i}}{\sum}-\log\frac{\exp\left(z_{i}\cdot z_{p}/\tau\right)}{\underset{b\in\mathcal{B}_{i}\setminus x_{i}}{\sum}\exp\left(z_{i}\cdot z_{b}/\tau\right)},\label{eq:SCLloss}}
\end{equation}
where $\tau$ is a temperature scaling constant and both the numerator and denominator compute inner products between pairs of offers embeddings. The former does it in-between positive offers, while the latter normalizes it using all offers in the batch $\mathcal{B}_{i}$ except the current $i$ offer. 
Minimizing Eq.~\ref{eq:SCLloss} implies adjusting $\theta$ and $\phi$ to pull together the feature representations $z_{i}$ and $z_{p}$ of positive product offers, while pushing apart $z_{i}$ from all negative product offers in $\mathcal{B}_{i}\setminus\mathcal{P}_{i}$. 
The gradient analysis of $\mathcal{L}_{SCL}$ in \cite{2021_NeurIPS_SupContLearn} reveals this training objective focuses on hard positives/negatives rather than easy ones. 
Introducing the blocking information via $\mathcal{P}_{i}$ and $\mathcal{N}_{i}$ promotes data-to-data relations between hard positive/negative samples.

\subsection{Product matching classification}
\label{sec3_2_finetuning}
\begin{figure}[htbp]
    \centering
    \includegraphics{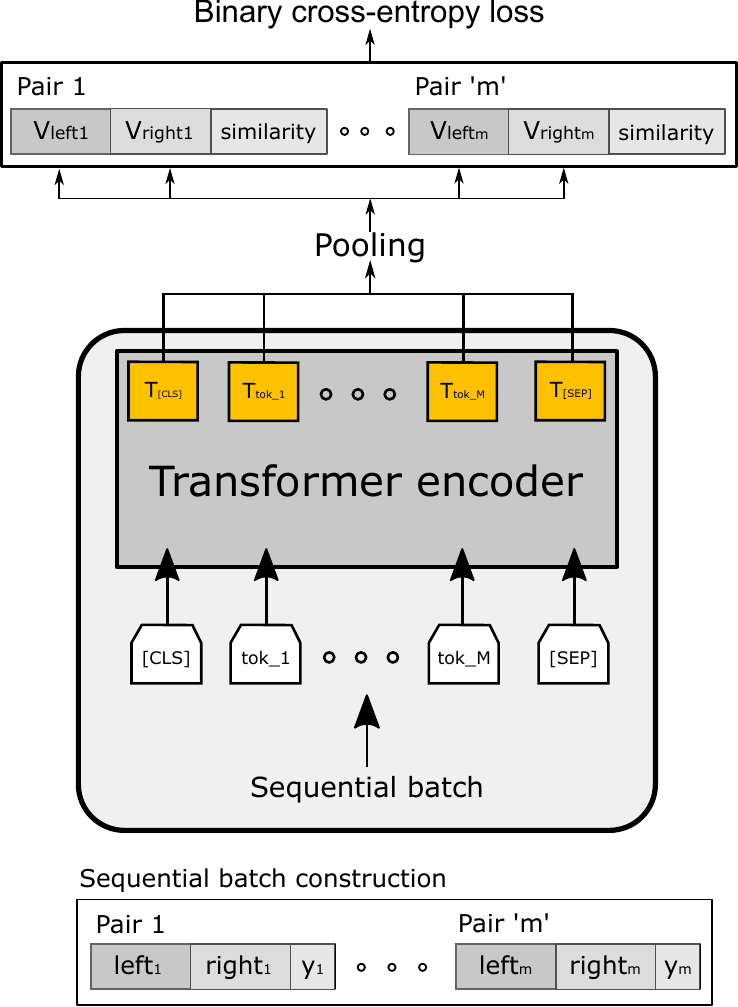}
    \caption{Diagram of the proposed downstream strategy and the sequential pair-based batch construction.}
    \label{fig:diagram_finetune}
\end{figure}

This last stage computes binary matching predictions for offer pairs by learning a linear layer on top of the frozen pre-trained transformer backbone. Figure \ref{fig:diagram_finetune} shows an overview of the product matching classification. The classification head receives the concatenation of the sentence embeddings from the two offers to match, $v_{i}$ and $v_{j}$, together with the concatenation of the Euclidean distance and cosine similarity of $v{i}$ and $v_{j}$. Then, the classifier layer input is $({v_{i}, v_{j}, |v_{i}-v_{j}|, v_{i}*v_{j}})$. For simplicity the two last terms are referred as \emph{similarity} in Figure~\ref{fig:diagram_finetune}. Note that the concatenation of this type of information is a standard practice in other works \cite{2020_EMNLP_SentenceEmbForBERT, 2022_arXiv_SupConLossPM} to help converge to top matching performance, something we have also observed in our experiments. Furthermore, the previously mentioned strategy assumes that $v_{i}$ goes before $v_{j}$ in the concatenation classifier, while there is no reason not to use the reversed order. Therefore, we also add the reversed concatenation order $({v_{j}, v_{i}, |v_{i}-v_{j}|, v_{i}*v_{j}})$ and average the logits of both orders, which we experimentally saw to slightly help in boosting matching performance.

%% file: sec4_experiments.tex
In this section we describe first, the datasets used to validate our approach, then the experimental setup and the results obtained compared to several related works.
Finally, we present an ablation study on the influence of the number of positives and negatives in the construction of the blocking within the batch.

\begin{table}[htbp]
  \centering
  \caption{Datasets general statistics: train split.}
  \label{tab:datasets_train}
  \begin{tabular}{llp{0.05\textwidth}p{0.05\textwidth}p{0.05\textwidth}p{0.05\textwidth}}
    \toprule
    Dataset & Size & \#\small{pos. pairs} &\#\small{neg. pairs} & \#\small{offers} & \#\small{prods.}\\
    \midrule
    \small{Abt-Buy} & def. & $822$ & $6,837$ & $2,112$ & $1,084$\\
    \small{Amzn-Goog.} & def. & $933$ & $8,234$ & $3,445$ & $2,279$\\
    \midrule
    \multirow{4}{*}{\small{WDC-Comp.}}
 
            & small & $722$ & $2,112$ & $2,790$ & $745$\\
            & med. & $1,762$ & $6,332$ & $3,846$ & $745$\\
            & large & $6,146$ & $27,213$ & $4,238$ & $745$\\
            & xlarge & $9,690$ & $58,771$ & $4,307$ & $745$\\
  \bottomrule
\end{tabular}
\end{table}

\begin{table}[htbp]
  \centering
  \caption{Datasets general statistics: test split.}
  \label{tab:datasets_test}
  \begin{tabular}{lp{0.08\textwidth}p{0.08\textwidth}p{0.08\textwidth}}
    \toprule
    Dataset & \#pos. pairs&\#neg. pairs & \#prods.\\
    \midrule
    Abt-Buy  & $206$ & $1,710$ & $921$\\
    Amzn-Goog.  & $234$ & $2,059$ & $1,962$\\
    \midrule
    WDC-Comp. & $300$ & $800$ & $745$\\
  \bottomrule
\end{tabular}
\end{table}

\begin{table}[htbp]
  \centering
  \caption{Datasets blocking statistics: train set.}
  \label{tab:train_blocking_stats}
  \begin{tabular}{llp{0.08\textwidth}p{0.08\textwidth}p{0.08\textwidth}}
    \toprule
     Dataset & Size & Average block. size & Average pos/block. & Average neg/block.\\
    \midrule
    Abt-Buy  & def. & $11.9 \pm 13.4$ & $1.2 \pm 1.0$ & $10.6 \pm 13.1$ \\
    Amzn-Goog. & def.  & $15.0 \pm 125.5$ & $2.8 \pm 41.1$ & $12.1 \pm 84.8$ \\
    \midrule
    \multirow{4}{*}{WDC-Comp.}
 
            & small  & $6.1 \pm 2.6$ & $1.5 \pm 0.8$ & $4.5 \pm 2.4$\\
            & med.  & $17.3 \pm 6.7$ & $3.7 \pm 1.8$ & $13.5 \pm 6.1$\\
            & large  & $71.6 \pm 30.3$ & $13.1 \pm 9.4$ & $58.4 \pm 25.3$\\
            & xlarge  & $146.8 \pm 92.4$ & $20.7 \pm 22.7$  & $126.0 \pm 77.7$\\
  \bottomrule
\end{tabular}
\end{table}

\begin{table*}[htp]
  \caption{F1-score results on the test set of each dataset. The symbol $^\clubsuit$ denotes that we run our implementation of the method. The results marked with $^*$ where taken from~\cite{2022_arXiv_SupConLossPM}. The $^\dag$ symbol indicates the rows used to aggregate results. Bold denotes the best performing results and underlined the second best.}
  \label{tab:general_res}
  \begin{tabular}{lllllllll|l}
    \toprule
    \multirow{2}{*}{Backbone} & \multirow{2}{*}{\#param.} & \multirow{2}{*}{Approach} & \multirow{2}{*}{Abt-Buy} & \multirow{2}{*}{Amzn-Goog.} & \multicolumn{4}{c}{WDC-Computers} & \multirow{2}{*}{\textbf{Avg.}}\\ 
    &&&&& small & medium & large & xlarge & \\
    \midrule
    \multirow{3}{*}{RoBERTa} & \multirow{3}{*}{110M} & Ditto baseline~\cite{2020_VLDB_Ditto} & $91.05^*$ & $65.92$ & $86.37^*$ & $91.90^*$ & $94.68^*$ & $94.73^*$ & $88.80^*$\\
    & &Ditto~\cite{2020_VLDB_Ditto} & $89.33$ & $75.58$ & - & - & - & - & $86.90^\dag$ \\
    & &Vanilla SCL (w/ tricks)~\cite{2022_arXiv_SupConLossPM} & $\mathbf{93.70}$ & $79.28$ & $\mathbf{93.18}$ & $\mathbf{97.66}$ & $\mathbf{98.16}$ & $\mathbf{98.33}$ & $\underline{93.38}$\\
    \midrule
    BERT-base & 110M & JointBERT~\cite{2021_VLDB_jointBERT} & - & - & $77.55$ & $88.82$ & $96.90$ & $97.49$ & -\\
    \midrule
    DistilBERT & 66M & Ditto~\cite{2020_VLDB_Ditto} & - & - & $80.76$ & $88.62$ & $91.70$ & $95.45$ & $86.90^\dag$ \\
    \midrule
    \multirow{3}{*}{BERT-med} & \multirow{3}{*}{35.3M} & Fine-tuned LM & $10.43$ & $21.09$& $10.03$& $26.86$ & $70.73$ & $79.16$ & 36.38 \\
    & & Vanilla SCL (w/o tricks)~\cite{2022_arXiv_SupConLossPM}$^\clubsuit$ & $90.95$ & $\underline{80.69}$ & $87.26$ & $92.63$ & $94.67$ & $94.67$ & $90.14$\\
    & & Blocking SCL (Ours) & $\underline{93.30}$ & $\mathbf{86.61}$ & $\underline{90.49}$ & $\underline{96.97}$ & $\underline{98.00}$ & $\underline{97.83}$ & $\mathbf{93.86}$\\
    \midrule
    Bi-LSTM & - & DeepMatcher & $62.8$ & $70.70$ & $61.22$ & $69.85$ & $84.32$ & $88.95$ & $72.97$\\
  \bottomrule
\end{tabular}
\end{table*}

\subsection{Datasets}
We conduct all our experiments using $6$ public datasets that are commonly used in the field of e-commerce and product matching: Amazon-Google \cite{2018_SIGMOD_DeepMatcher}, Abt-Buy \cite{2018_SIGMOD_DeepMatcher} and WDC \cite{2019_WWWC_WDCdataset} (all variants of the computers subset, \ie~\emph{small}, \emph{medium}, \emph{large} and \emph{xlarge}).
Tables~\ref{tab:datasets_train} and~\ref{tab:datasets_test} shows general statistics of the train and test splits.
Unlike other methods~\cite{2022_arXiv_SupConLossPM}, we only use the \emph{title} as an input feature (or equivalently the shortest textual information used to describe the product offer), which greatly simplifies the training, \ie~fewer memory requirements, and computations, while achieving better performance.

Additionally, we inspect the statistics related to the blocking information as shown in Table~\ref{tab:train_blocking_stats}. In the table, we present three main columns: \emph{Avg. block. size}, which refers to the number of product textual descriptions that are candidates to be a match for each query product description. 
The \emph{Avg. pos/block} representing the match descriptions out of the candidates and, the \emph{Avg. neg/block} for the non-match descriptions. 
Note that each query product might have different blocking sizes as well as the number of positive and negative descriptions, thus we present the average statistics across all query products.
We can observe significant differences in these three aspects between datasets, for example, a ${\sim30\times}$ increase in the number of negatives per blocking in the \emph{xlarge} version compared to the \emph{small} version. Note that the specific characteristics of the blocking in each dataset might yield different optimal configurations of our method. For instance, for datasets with more number of negatives in the candidates' list we might want to select more negatives per blocking to increase variability in the batch.

\subsection{Experimental setup}
We use a pre-trained BERT medium (BERT-med)\footnote{\url{https://huggingface.co/google/bert_uncased_L-6_H-512_A-8}} as the backbone ($6$ layers, $8$ attention heads, and $512$ hidden dimension size) in all the experiments and a cosine annealing scheduler for the learning rate with $5\%$ of warm-up period and maximum value of $5 \cdot 10^{-5}$.

For the supervised contrastive learning pre-training, we use a projection head composed of one linear layer of dimension $512$ followed by \emph{gelu} activation, dropout with probability $0.1$, \emph{LayerNorm}, and a final linear projection to a low-dimensional representation of $256$. 
We train for $300$ epochs in Abt-Buy, Amazon-Google, and WDC-small-computers, $400$ epochs in WDC-medium-computers, $600$ epochs in WDC-large-computers, and $800$ epochs in WDC-xlarge-computers. 
Regarding the hyper-parameters related to the blocking, for the smaller datasets Abt-Buy, Amazon-Google and WDC-small-computers we found the optimal configuration with $1$ positive and $16$ negatives while, for the rest we used $2$ positives and $16$ negatives. In general we have observed stability of the models on the blocking parameters, however, a more rigorous analysis will be required to understand the relation between dataset and the selection of these hyper-parameters.
After training, we select the model checkpoint with the lowest validation loss for the subsequent fine-tuning.

To estimate a product matching prediction, we replace the projection head learned during pre-training with a linear layer that maps the $512$ dimensional sentence embedding from the Transformer backbone to the matching/not-matching space. 
We train for $50$ epochs freezing the backbone and use early-stopping if the validation loss does not decrease in $10$ epochs. 
For evaluation, we use the checkpoint with the highest F1-score in validation.

\subsection{Results}

Table~\ref{tab:general_res} reports the resulting F1 scores of our experiments along with top-performing methods in $6$ commonly used public datasets. 
The compared methods are \textsc{Ditto}~\cite{2020_VLDB_Ditto}, Vanilla-SCL~\cite{2022_arXiv_SupConLossPM}, JointBERT~\cite{2021_VLDB_jointBERT} and DeepMatcher~\cite{2018_SIGMOD_DeepMatcher}. 
We name the recent method in \cite{2022_arXiv_SupConLossPM} as Vanilla-SCL given that it performs random sampling of negatives following  \cite{2021_NeurIPS_SupContLearn}, as opposed to our Block-SCL method that uses negative mining from blocking information. 
For Vanilla-SCL we report the results from \cite{2022_arXiv_SupConLossPM} using RoBERTa-base backbone, which we name Vanilla-SCL with tricks \emph{(w/ tricks)} due to using data augmentation and serialization of multiple input features (title, description and additional information) with additional tagging of the data field and value using special tokens [COL] and [VAL]. 
We further reproduce Vanilla-SCL without tricks \emph{(w/o tricks)}, \ie~no data augmentation and using only the title with no tagging strategy, and run it ourselves using BERT-med backbone to enable fair comparison with our Block-SCL.

Our method surpasses by a large margin (${\uparrow7.3}$) the results of the previous top-performing method~\cite{2022_arXiv_SupConLossPM} in the Amazon-Google dataset. 
This dataset is the less saturated one in terms of performance, giving sufficient room for improvement, unlike the other $5$ datasets, where related work performance ranges from $93.16$ up to $98.1$ F1-score. 
In the remaining datasets, our method achieves comparable results to~\cite{2022_arXiv_SupConLossPM} using a model $3\times$ smaller and a more modest training strategy \eg~smaller batch-sizes and input sequences. 
Regarding efficiency, we conducted a preliminary evaluation of training times simulating the configuration proposed by~\cite{2022_arXiv_SupConLossPM}, the results are presented in Table~\ref{tab:computational_time}. 
Our configuration is significantly more efficient, having a $4\times$ and $2\times$ speed up with BERT-med and RoBERTa, respectively. 
These results suggest that building complex batches with hard negatives brings benefits during training, leading the convergence of the model to a more optimal minimum, without requiring large architectures.

The fine-tuned LM approach in Table~\ref{tab:general_res} refers to using the pre-trained BERT-med backbone from a general domain and entirely fine-tuning it on the downstream task of product matching following the approach described in section \ref{sec3_2_finetuning}. 
This strategy performs poorly, thus demonstrating that the pre-trained model is key to learning meaningful sentence embeddings for product matching.

Due to computational restrictions, we were not able to run our method with RoBERTa-base and the same configuration proposed in~\cite{2022_arXiv_SupConLossPM}. 
However, we should expect some improvement when using RoBERTa-base ($110$M) compared to BERT-med ($35.3$M), as bigger architectures, trained on larger datasets and batch sizes usually come with performance gains in supervised contrastive learning \cite{2021_NeurIPS_SupContLearn}. 
Moreover, data augmentation in NLP is known to be hard, so we do not apply it given that using labels, \ie~product identifiers in our scenario, reduces the importance of data augmentation in contrastive learning \cite{2020_NeurIPS_GoodViews}.

\subsection{Ablation studies}
A key component of our approach is the construction of the batches, as described in \ref{sec3_1_2_batch_construction}. 
In this stage, we sample a certain number of positive and negative pairs from each blocking. 
However, it is unclear how many positives and negatives we should select. 
Table~\ref{tab:ablation_res} shows the results of not using the blocking in the batches, as in Vanilla-SCL, and presents Block-SCL performance when varying the number of positives and negatives. 
Regarding the impact of including hard-negatives in the batch we report an improvement of $2.8$ points with respect to not using hard-negatives, \ie~random sampling. 
Additionally, we observe a clear tendency for improvement as the number of negatives increases in the batch. 
These results verify the effectiveness of our approach and our initial hypothesis that the use of hard negatives from the blocking examples within the same batch should lead the model convergence to a more optimal solution. 
Finally, we do not observe any impact of the number of positive samples chosen from each blocking, which 
suggests that positives within a blocking are relatively redundant and provide a very similar training signal.
Nevertheless, further analysis need to be conducted to gain a more solid understanding of these parameters influence.

The experiments in this section are run on the WDC Computers medium dataset using the same configuration except for the hyper-parameters stated in the Table \ref{tab:ablation_res}.
Concretely, we used the BERT-med architecture, the training was optimized with AdamW using a warm-up of $5\%$ followed by a cosine annealing learning rate decay strategy. 
For the pre-training stage we used a batch size of $256$, and trained for $200$ epochs with an initial learning rate of $10^{-4}$. 
In the product matching classification, we used a batch size of $64$ during $50$ epochs with an initial learning rate of $5\cdot 10^{-5}$. 
Finally, we report the average of $3$ runs along with the standard deviation, which demonstrates the stability of Block-SCL.

\begin{table}[htbp]
  \caption{Influence of the number of positives/negatives in WDC-medium. We report the average and standard deviation of the F1-score values after running $3$ repetitions of Block-SCL with different seeds.}
  \label{tab:ablation_res}
  \begin{tabular}{cccc}
    \toprule
     Blocking & \#pos & \#neg & F1\\
    \midrule
    No & $1$ & $0$ & $93.15 \pm 0.52$ \\
    \midrule
    \multirow{4}{*} {Yes} & $1$ &8 & $95.96 \pm 0.20$ \\ 
    & $2$ & $8$ & $95.91 \pm 0.46$ \\ 
    & $3$ & $8$ & $96.32 \pm 0.15$ \\ 
    & $6$ & $8$ & $96.17 \pm 0.13$ \\ 
    \midrule
    \multirow{4}{*} {Yes} & $1$ & $1$ & $93.88 \pm 0.31$ \\ 
    & $1$ & $2$ & $94.89 \pm 0.33$ \\ 
    & $1$ & $4$ & $95.01 \pm 0.52$ \\ 
    & $1$ & $16$ & $96.13 \pm 0.22$ \\ 
    \midrule
    Yes & $2$ & $16$ & $96.53 \pm 0.15$ \\ 
  \bottomrule
\end{tabular}
\end{table}

\begin{table}[htbp]
  \caption{Computational time between different settings.}
  \label{tab:computational_time}
  \begin{tabular}{llcc}
    \toprule
     Approach & Backbone & it/s & epoch/min\\
    \midrule
    \multirow{2}{*} {Block-SCL (Ours)} & BERT-med & $9.30$ & $2$\\
    & RoBERTa & $2.84$ & $0.80$ \\
    \midrule
    \multirow{2}{*} {Vanilla SCL~\cite{2022_arXiv_SupConLossPM}} & BERT-med & $2.31$ & $0.66$ \\
    & RoBERTa & $1.25$ & $0.18$ \\
  \bottomrule
\end{tabular}
\end{table}

\hspace{1cm}



%% file: sec5_conclusions.tex
In this paper, we tackle the problem of product matching in e-commerce and focus on learning robust representation of products using supervised contrastive learning as training loss. This learning objective based on metric learning suits the product matching task, which essentially measures similarities between textual product offers. 
Building on top of this intuition, we incorporate the blocking information within the optimization loop in a way that every batch randomly samples several disjoint blockings (unique product ids). 
We empirically demonstrate that negative samples within the blocking introduce a strong signal in the supervised contrastive loss formulation, obtaining high-quality product embeddings.

In the experimental section, we demonstrate that our proposed Block-SCL method is able to learn more discriminative product embeddings by measuring the product matching performance based on learning a product matching classifier on top of those embeddings. We outperform most related work methods by a large margin and clearly improve over Vanilla-SCL \cite{2022_arXiv_SupConLossPM} when compared in a fair setup without tricks. Nevertheless, considering Vanilla-SCL with tricks (data serialization of multiple product fields and data augmentation) and a larger transformer backbone, we perform on par, while being more than 4 times faster.

We leave for future work an in-depth analysis on the impact of the number of blockings in a batch, and positives and negatives within each blocking, as well as experimenting with larger backbones and tricks whose impact is not clear in previous work.